\documentclass{article}

\usepackage[final]{ewrl_2025}

\usepackage{natbib}
\setcitestyle{authoryear,open={(},close={)}}
\usepackage{hyperref}
\hypersetup{
colorlinks   = true,    
urlcolor     = black,   
linkcolor    = black,   
citecolor    = black 
}
\usepackage[utf8]{inputenc}
\usepackage[T1]{fontenc}  
\usepackage{hyperref}       
\usepackage{url}           
\usepackage{booktabs}     
\usepackage{amsfonts}       
\usepackage{nicefrac}     
\usepackage{microtype}      
\usepackage{xcolor}         
\usepackage{amsthm}
\usepackage{amsmath}
\usepackage{algorithm}
\usepackage{algpseudocode}

\title{Locally Differentially Private Thresholding Bandits}

\author{
  Annalisa Barbara \\
  Bocconi University \\
  \texttt{annalisa.barbara@studbocconi.it} \\
  \And
  Joseph Lazzaro \\
  Imperial College London \\
  \texttt{joseph.lazzaro18@imperial.ac.uk} \\
  \AND
  Ciara Pike-Burke \\
  Imperial College London \\
  \texttt{c.pike-burke@imperial.ac.uk} \\
}

\newtheorem{definition}{Definition}[section]
\newtheorem{theorem}{Theorem}[section]  
\newtheorem{lemma}[theorem]{Lemma} 
\newcommand{\KL}[2]{D_{\mathrm{KL}}\left(#1 \, \| \, #2\right)}
\newcommand{\Prob}{\mathbb{P}}
\newcommand{\E}[2]{\mathbb{E}_{#1}\left[#2\right]}
\newcommand{\DPconst}{2\min\{4, e^{2\epsilon}\}(e^\epsilon - 1)^2}

\begin{document}
\maketitle

\begin{abstract}
This work investigates the impact of ensuring local differential privacy in the thresholding bandit problem. We consider both the fixed budget and fixed confidence settings. We propose methods that utilize private responses, obtained through a Bernoulli-based differentially private mechanism, to identify arms with expected rewards exceeding a predefined threshold.
We show that this procedure provides strong privacy guarantees and derive theoretical performance bounds on the proposed algorithms. Additionally, we present general lower bounds that characterize the additional loss incurred by any differentially private mechanism, and show that the presented algorithms match these lower bounds up to poly-logarithmic factors. Our results provide valuable insights into privacy-preserving decision-making frameworks in bandit problems.
\end{abstract}

\section{Introduction}
\label{sec:introduction}
Multi-armed bandit (MAB) problems \cite[e.g.,][]{Berry1984BanditPS} represent a foundational framework for sequential and stochastic decision-making processes, where information is gathered and utilized iteratively. These problems have a broad spectrum of practical applications in settings where decisions must be made under uncertainty such as optimizing clinical trials \citep[e.g.,][]{villar15} and enhancing adaptive streaming systems \citep[e.g.,][]{pmlr-v139-jin21a}. In this work, we focus on a pure-exploration variant of the stochastic MAB problem, where each arm has a bounded reward drawn from an unknown distribution. The objective is to identify all arms whose expected rewards exceed a specified threshold. This problem, known as the thresholding bandit problem, has been extensively investigated \cite[e.g.,][]{locatelli16, Chen14, Chelsire20}, and optimal algorithms have been established. This setting also has practical relevance, for example, in dose-finding applications in healthcare, where identifying treatments above an efficacy threshold is critical \citep{garivier2017thresholding}.

In recent years, there has been a growing interest in integrating privacy-preserving mechanisms into bandit algorithms \cite[e.g.,][]{basu2019privacy, Ren20}, driven by the increasing demand for safeguarding sensitive information in decision-making processes. This is particularly relevant in medical applications, for instance, when bandit algorithms are used for adaptive treatment decisions and patient data must remain private.
Differential privacy (DP) has emerged as a rigorous mathematical framework for quantifying privacy and is widely regarded as a standard in this domain \citep{dwork2014algorithmic}. However, incorporating DP into MAB settings introduces significant challenges, as privacy constraints inherently limit the quality of information that can be used for decision-making.
    
This paper explores the thresholding bandit problem under Local Differential Privacy (LDP) guarantees. We adapt existing algorithms designed for the non-private case to meet privacy requirements in the thresholding bandits problem and quantify the impact of ensuring privacy. We consider the two standard settings:

\begin{itemize}
    \item \textbf{Fixed budget setting:} Given a limited budget on the total number of actions we can make, $T$, the goal is to identify arms with expected rewards exceeding a given threshold. We demonstrate how the inclusion of privacy constraints influences the probability of correctly identifying the desired set of arms after $T$ rounds.
    \item \textbf{Fixed confidence setting:} For a given confidence level $\delta$, the objective is to minimize the number of interactions needed to confidently identify the correct set of arms with probability greater than $1-\delta$. We analyze the impact of privacy guarantees on the expected sample complexity, showing an increase in the number of interactions needed to meet the confidence criteria.
\end{itemize}

This work contributes to the understanding of how privacy constraints influence the performance of thresholding bandit algorithms, providing insights for designing effective privacy-preserving decision-making systems. Additionally, we derive lower bounds for both the fixed budget and fixed confidence scenarios and show that the algorithms presented match these bounds up to poly-logarithmic factors. In this work, we further extend the understanding of integrating privacy guarantees into bandit settings. While prior work has analyzed the best arm identification problem under privacy constraints, we propose an analysis for an equally important setting: the thresholding bandit problem, where the goal is to identify all arms with mean rewards above a given threshold.

The remainder of this paper is organized as follows. Section~\ref{sec:setting} introduces the thresholding bandit setting, the notion of differential privacy, gives one example of a differentially private mechanism (the Bernoulli mechanism), and defines the concept of locally differentially private thresholding bandits. Section~\ref{sec:FB_alg} introduces a locally differentially private algorithm for the fixed budget case using the Bernoulli mechanism.  
Section~\ref{sec:FB_lb} provides general lower bounds for any locally differentially private bandit algorithm in the fixed budget setting. 
Section~\ref{sec:FC_alg} discusses a locally differentially private algorithm for the fixed confidence case, also leveraging the Bernoulli mechanism.  
Finally, Section~\ref{sec:FC_lb} presents general lower bounds for any locally differentially private bandit algorithm in the fixed confidence setting. We conclude in Section~\ref{sec:conclusion}.

\section{Preliminaries}
\label{sec:setting}

\subsection{Related Work}

\textbf{Pure Exploration} The classic bandit pure exploration problem is Best Arm Identification (BAI) in which the learner's objective is to confidently identify the arm with the highest mean reward using the fewest number of actions. This has been done in both the fixed-confidence \citep[e.g.][]{kaufmann2016complexity} and fixed-budget settings \citep[e.g.][]{fixed_budget_BAI_2010}, as outlined in the introduction. Another direction of pure exploration research is thresholding bandits, which has also been studied in both the fixed-confidence and fixed-budget settings \cite[e.g.,][]{locatelli16, Chen14, Chelsire20,garivier2017thresholding}, and will be the focus of this paper. In thresholding bandits, rather than just returning the arm with the largest mean reward as in BAI, the learner's objective is to return all arms with mean reward larger than a given threshold. See Section \ref{sec:setting_threshold} for a formal definition and introduction. While the thresholding bandit problem has been well understood in the standard bandit feedback setting, it has not been studied under privacy constraints.

\textbf{Privacy in Bandits} For applications in which rewards themselves could represent sensitive user information, different privacy frameworks have been proposed. The first and strongest form of privacy studied in bandits has been Local Differential Privacy (LDP), in which the rewards are never directly observed by the learning algorithm and will first need to be passed through a privacy mechanism \citep{Ren20, han2021generalized}. This is the setting we will study in this paper, see Section~\ref{sub:DP}  for more details. A weaker form of privacy in bandits is Global Differential Privacy (GDP), in which the learning algorithm is trusted to directly observe the rewards, however the sequence of decisions made by the algorithm should not provide untrustworthy third parties information about any individual reward during the learning process \citep[e.g.][]{azize2022privacy,azize2024differentially}. Finally, in contextual bandit settings, Joint Differential Privacy (JDP) has been proposed \citep[e.g.][]{shariff2018differentially,pavlovicdifferentially}. Here the learner is restricted to keep both the rewards \emph{and} the contexts private from potentially untrustworthy third parties. While lots of work has been done studying privacy in bandits for various regret minimisation problems \citep[e.g.][]{Ren20}, less work has been done to study private pure exploration problems. Private BAI has been considered \citep[e.g.][]{azize2024differentially}, however privacy has not been studied with Thresholding Bandits. Hence, this paper fills this gap in the private bandits literature.

\subsection{Thresholding Bandits}\label{sec:setting_threshold}
Multi-armed bandit (MAB) problems model sequential decision-making under uncertainty, where an agent interacts with a set of $K$ arms over multiple rounds, collecting rewards while learning about the underlying reward distributions. In each round $t$, the agent selects an arm $i \in \mathcal{A} = \{1, 2, \dots, K\}$ and observes a stochastic reward $X_i(t) \in [0,1]$, which is drawn from an unknown distribution with mean $\mu_i = \mathbb{E}[X_i(t)]$. The objective in thresholding bandit problems is to identify arms whose mean rewards exceed a given threshold $\tau>0$, subject to either budgetary constraints (fixed budget setting) or confidence guarantees (fixed confidence setting).

\paragraph{Fixed Budget Thresholding Bandits}

In the fixed budget thresholding bandit problem \citep{locatelli16}, the goal is to identify a subset of arms from a set \( \mathcal{A} = \{1, 2, \dots, K\} \) whose mean rewards exceed a prespecified threshold, up to a tolerance factor $\zeta$. Formally, let each arm \( i \) in \( \mathcal{A} \) have a reward that is distributed within \([0,1]\) and an unknown mean reward \( \mu_i \). Let \( \tau \) denote the desired threshold, and let \( \Delta_i = |\tau - \mu_i| + \zeta\) represent the deviation of the mean reward of arm \( i \) from the threshold plus some tolerance $\zeta \geq 0$. The objective is to return \(\mathcal{S}_\tau \subseteq \mathcal{A} \) where:
\begin{equation}
    \mathcal{S}_\tau = \{ i \in \mathcal{A} : \mu_i \geq \tau + \zeta \}
\end{equation}
In other words, the set $\mathcal{S}_\tau$ contains arms whose mean rewards are greater than or equal to the threshold \(\tau\), increased by a margin of \(\zeta\). The parameter $\zeta$ introduces some tolerance in the classification of arms above the threshold. Specifically, we allow a margin of $2\zeta$ around the threshold within which arms may be misclassified, but we require that all arms whose means differ from the threshold by more than $\zeta$ are classified correctly.

The learner has a fixed budget of $T$ rounds during which they may sequentially choose arms to play. Their goal is to return a set of arms whose mean is above the threshold $\tau$. 
Let \( \hat{\mathcal{S}}_\tau \) be the set of arms returned after \( T \) arm pulls and  \( \hat{\mathcal{S}}^C_\tau \) be its complement. The classification error is defined as:
\begin{equation} \label{eq:classification_error}
\mathcal{L}(T) = \mathbb{I} \left\{ \exists i \in \hat{\mathcal{S}}_\tau : \mu_i \leq \tau - \zeta \lor \exists i \in \hat{\mathcal{S}}^C_\tau : \mu_i > \tau + \zeta  \right\}. 
\end{equation}
This means that the loss will be equal to 1 if any arm with mean above $\tau + \zeta$ is mistakenly excluded from $\hat{\mathcal{S}}_\tau$, or if any arm with mean below $\tau - \zeta$ is mistakenly excluded from $\hat{\mathcal{S}}_\tau^C$.
 The goal is to minimize the expected loss after $T$ rounds.

\paragraph{Fixed Confidence Thresholding Bandits}

In the fixed confidence thresholding bandit problem \citep{Chen14}, the goal is to identify a subset of arms whose mean is above the desired threshold $\tau$ with a confidence level of at least \( 1 - \delta \). 

Let \( \Delta_i = |\tau - \mu_i| \) represent the deviation from the threshold. The objective is to find the set of arms with mean rewards exceeding the threshold $\tau$, \(\mathcal{S}_\tau \subseteq \mathcal{A} \), where:
\begin{equation}
    \mathcal{S}_\tau = \{ i \in \mathcal{A} : \mu_i \geq \tau\}.
\end{equation}
The only difference compared to the fixed budget setting is that in the fixed confidence case, $\zeta$ is not used (i.e., $\zeta = 0$). As a result, the definitions of $\Delta_i$ differ slightly between the two settings. In line with previous works, we include a tolerance factor $\zeta$ in the fixed budget case to account for the limited number of samples, since the number of exploration rounds is bounded by $T$. On the other hand, in the fixed confidence setting, we are interested in the number of samples required to ensure that the returned set of arms is correct with probability $1-\delta$.

The objective is to minimize the number of interactions required to return a set $\hat{\mathcal{S}}_\tau$ that matches $\mathcal{S}_\tau$ with probability greater than $1-\delta$. Specifically, we want to guarantee that the probability of an error in classification is bounded by \( \delta \), i.e.
\[
\mathbb{P}[\hat{\mathcal{S}}_\tau \neq \{ i \in \mathcal{A} : \mu_i \geq \tau \}] \leq \delta.
\]

In this setting, the algorithm does not have a predetermined budget of arm pulls. Instead, it continues pulling arms until it believes the output set is correct with probability at least $1-\delta$, at which point the algorithm stops. Here the total expected number of pulls $T$ required before stopping is the parameter we aim to optimize.\\

\subsection{Differential Privacy}
\label{sub:DP}
Differential privacy guarantees that the output of an algorithm remains nearly indistinguishable when applied to two datasets differing in only one individual’s data \citep{dwork2014algorithmic}. This ensures that the participation of any individual has a negligible impact on the algorithm's output, effectively preventing leakage of sensitive information about that individual.

Formally, a mechanism \( \mathcal{M} \) is \( \epsilon \)-differentially private if, for any two datasets \( D \) and \( D' \) differing in one entry (i.e. they are identical except for a single individual's data being changed), and for any output \( S \subseteq \text{Range}(\mathcal{M}) \), it holds that:
\[
\Pr[\mathcal{M}(D) \in S] \leq e^\epsilon \cdot \Pr[\mathcal{M}(D') \in S],
\]
where \( \epsilon > 0 \) is the parameter controlling the level of privacy. This definition states that that for any two datasets differing in only one entry, their statistical properties are almost the same and so it is very difficult to infer the presence of a specific individual in a given dataset. Smaller values of $\epsilon$ guarantee high levels of privacy, in fact when $\epsilon=0$ the above becomes equivalent to requiring
\[
\Pr[\mathcal{M}(D) \in S] = \Pr[\mathcal{M}(D') \in S].
\]
On the other hand, when $\epsilon \rightarrow \infty$ no privacy is guaranteed.

\subsubsection{Bernoulli Mechanism}
\label{sub:Bern_mech}
Among the numerous studies on differential privacy mechanisms \citep{kasiviswanathan2011can, duchi2013local, kairouz2016extremal}, one of the most widely adopted methods for ensuring privacy involves incorporating randomization into users' data or responses. In this paper, we consider a specific mechanism, namely the Bernoulli differential private mechanism. This mechanism has previously been studied in other bandit settings, for example in \citet{Ren20} for locally differentially private bandit algorithms in regret minimization problems. Here, we propose employing the same mechanism within the alternative thresholding bandit framework.

The Bernoulli mechanism, which will be utilized in both the fixed budget (Section~\ref{sec:FB_alg}) and fixed confidence (Section~\ref{sec:FC_alg}) settings, transforms a bounded reward in $[0,1]$ into a Bernoulli random variable that satisfies differential privacy guarantees.
This mechanism produces a binary output (0 or 1) that serves as a privacy-preserving approximation of the reward. Consequently, the agent does not have direct access to the true reward  of the arm; instead, they observe only the realization of a Bernoulli random variable.

Now, we formally define the Bernoulli mechanism as in \citep[Lemma 5]{Ren20}, which will be used in the following sections.

\begin{definition}
Given a reward \( r \in [0,1] \) to be privatized, the \emph{PrivBern($\epsilon$) Mechanism} returns a binary value sampled from a Bernoulli distribution. Specifically, the mechanism:  

\begin{enumerate}
    \item Receives a reward \( r \in [0,1] \).
    \item Outputs \( B(r) \), where \( B(r) \) is an independent sample from:
    \[
    \text{Bernoulli} \left( \frac{r e^\epsilon + (1 - r)}{1 + e^\epsilon} \right).
    \]
\end{enumerate}
\end{definition}

\begin{lemma}\label{lem:privbern}
The PrivBern($\epsilon$) Mechanism is an $\epsilon$-DP mechanism that takes as input a reward in $[0,1]$ from arm $a$, and outputs a Bernoulli-distributed random variable with an expected value
\[
\mu_{a,\epsilon} := \frac{1}{2} + (2 \mu_a - 1) \cdot \frac{e^\epsilon - 1}{2(e^\epsilon + 1)},
\]
where $\mu_a$  denotes the true mean reward of arm $a$.
\end{lemma}
For completeness, we include the proof in Appendix~\ref{appendix:proof_privbern}, as given in \cite{Ren20}.

\subsubsection{Locally Differentially Private Thresholding Bandits}
\label{sub:LDPbandits}
A particularly relevant variant of differential privacy (DP) is Local Differential Privacy (LDP) \citep[e.g.][]{yang2024local}, which is commonly applied in scenarios where data is collected from multiple users without relying on a trusted central entity. Unlike traditional DP, which assumes there is a trusted curator who has access to the whole data set and applies noise to the aggregated data, LDP requires that individual data is locally randomized  before sharing. This ensures that raw data never leaves the user's device in its original form, significantly enhancing privacy in applications such as data collection for statistics, where users may wish to protect their sensitive personal information from being directly shared with a central entity. 

In the context of bandit algorithms, this requirement means that the agent does not directly observe the true reward when pulling an arm, but rather a privatized, noisy version of it. The amount of noise introduced depends on the desired level of privacy. As the privacy requirement increases, more noise will be added to the true (unobserved) reward, making it increasingly difficult for the agent to accurately estimate the true mean reward of each arm. We now give the definition of an $\epsilon$-LDP bandit algorithm \citep{pmlr-v83-gajane18a}.

\begin{definition}\label{def:LDPBandit}
A bandit algorithm is locally differentially private if, after each action, the algorithm observes only the output of an $\epsilon$-DP mechanism applied to the generated reward.
\end{definition}

Hence in this work on locally differentially private thresholding bandits, our objective is to identify the set of arms with means which exceed a threshold (as described in Section \ref{sec:setting_threshold}) while \emph{also} guaranteeing that our proposed algorithm is $\epsilon$-LDP. We emphasize that since the Bernoulli Mechanism described in Section~\ref{sub:Bern_mech} is an $\epsilon$-DP mechanism, any bandit algorithm that observes only the privatized rewards produced by this mechanism is, by definition, a locally differentially private bandit algorithm in the sense of Definition~\ref{def:LDPBandit}.

\subsection{Preliminary Notation}
We introduce some additional notation. Let $\tau > 0$ be the desired threshold, $\zeta$ the tolerance used when classifying arms, and for each arm $i \in [K]$, let $\mu_i$ denote the mean of arm $i$. We define a quantity $H$ that captures the difficulty of the problem as:

$$
H = \sum_i \left( |\mu_i - \tau| + \zeta \right)^{-2}.
$$

Note that in the fixed confidence setting, we set $\zeta = 0$.

We now introduce the following quantities:
\begin{align}
\mu_{i,\epsilon} &= \frac{1}{2} + \left( 2\mu_i - 1 \right) \frac{e^\epsilon - 1}{2(e^\epsilon + 1)} \label{eq:mu_eps},\\
\tau_\epsilon &= \frac{1}{2} + \left( 2\tau - 1 \right) \frac{e^\epsilon - 1}{2(e^\epsilon + 1)},\\
\zeta_\epsilon &= \frac{e^\epsilon - 1}{e^\epsilon + 1} \zeta,\\
\Delta_{i,\epsilon} &= |\tau_\epsilon - \mu_{i,\epsilon}| + \zeta_\epsilon,\\
H_\epsilon &= \sum_i \left( |\mu_{i,\epsilon} - \tau_\epsilon| + \zeta_\epsilon \right)^{-2}.
\end{align}

\section{Fixed Budget Case}
\subsection{Algorithm for the Fixed Budget Case}
\label{sec:FB_alg}

In this section, we introduce the private version of the fixed budget thresholding bandit algorithm, building on the standard non-private formulation presented by \cite{locatelli16}. The key distinction from the non-private setting is that, in the private setting, the agent does not observe the true reward but instead interacts with privatized responses. By applying the $PrivBern(\epsilon)$ Mechanism to guarantee privacy (Lemma~\ref{lem:privbern}), we see that the agent is now interacting with Bernoulli random variables whose mean is \(\mu_{i,\epsilon}\), as defined in Equation~\eqref{eq:mu_eps}. 
Hence, to adapt the non-private algorithm to work with the privatized rewards, we need to carefully rescale the threshold.
We briefly outline our proposed algorithm and provide an upper bound for the expected loss.

Algorithm \ref{alg:FB_alg} begins by taking as input a threshold $\tau$, a tolerance $\zeta$, and a privacy parameter $\epsilon$. It starts by pulling each arm once and collecting private responses obtained from a $PrivBern(\epsilon)$ mechanism. For each arm $i \in [K]$, the empirical mean of the private responses is computed up to time $t$, denoted $\hat{\mu}_i^t$. At each round the algorithm selects an arm $I_t$ to pull by minimizing the term $$B_k(t) = \sqrt{T_k(t-1)}\hat{\Delta}_{k,\epsilon}(t-1),$$ where $T_k(t-1)$ is the number of pulls of arm $k$ up to time $t-1$ and $$\hat{\Delta}_{k,\epsilon}(t-1)= |\tau_\epsilon - \hat{\mu}_{k}^t| + \zeta_\epsilon$$ is the estimate of arm $k$'s private reward deviation from $\tau_\epsilon$
at time $t-1$, and $$\tau_\epsilon = \frac{1}{2} + \left( 2\tau - 1 \right) \frac{e^\epsilon - 1}{2(e^\epsilon + 1)}.$$ After selecting the arm, the algorithm pulls it, collects a private response from $PrivBern(\epsilon)$, and updates the empirical mean for the selected arm. 
The algorithm never observes the true reward from pulling the arm, only the privatized version. Let $\hat{\mu}_k(T) $ be the empirical mean of the private responses of arm $k$ up to time $T$. 
After $T$ rounds, the algorithm returns the set $$\hat{S}_\tau = \{k \in [K] : \hat{\mu}_k(T) > \tau_\epsilon\},$$ which contains the arms whose estimated rewards exceed the threshold $\tau_\epsilon$.

\begin{algorithm}[H]
\caption{Fixed budget thresholding LDP-bandit algorithm}\label{alg:FB_alg}
\begin{algorithmic}[1]

\State \textbf{Input:} $\tau$, $\zeta$ $\epsilon$ (threshold, tolerance, and DP-parameter)

\State Pull each arm once and receive private responses from $PrivBern(\epsilon)$

\State $t \gets K$

\State $\hat{\mu}_i^t \gets$ empirical mean of the private responses of arm $i$ till time $t$

\For{$t = K+1, \dots, T$}
    \State Pull arm $I_t \gets \arg\min_{k \leq K} B_k(t)$ 
    \Comment where $B_k(t) \gets \sqrt{T_k(t-1)}\hat{\Delta}_{k,\epsilon}(t-1)$
    \State Observe the private response from $PrivBern(\epsilon)$ and update the mean of the pulled arm
\EndFor

\State \Return $\hat{S}_\tau = \{k : \hat{\mu}_k(T) > \tau_\epsilon\}$ 

\end{algorithmic}
\end{algorithm}

Recalling the loss definition from Equation~\eqref{eq:classification_error}, we now establish an upper bound on the expected loss after \( T \) rounds of Algorithm~\ref{alg:FB_alg}, given by  

\[
\mathbb{E}\left[\mathcal{L}(T)\right] = \mathbb{P}\left\{ \exists i \in \hat{\mathcal{S}}_\tau : \mu_i \leq \tau - \zeta \lor \exists i \in \hat{\mathcal{S}}^C_\tau : \mu_i > \tau + \zeta  \right\},
\]
and show that Algorthm~\ref{alg:FB_alg} is $\epsilon$-LDP.

\begin{theorem}\label{thm:FB_alg}
Let $K > 0$ and  $T \geq 2K$. Assume that all the arms have rewards in $[0,1]$ with mean $\mu_k$. Let $\tau \in \mathbb{R}, \zeta \geq 0, \epsilon \geq 0$.
The expected loss of Algorithm \ref{alg:FB_alg} is upper bounded
by
\[
\mathbb{E}\left[\mathcal{L}(T)\right] \leq \exp\left(-\frac{T}{4H_\epsilon} + 2K \log\left(\log(T) + 1\right)\right),
\]
where
\[
H_\epsilon = \sum_{i=1}^K (|\mu_{i,\epsilon} - \tau_\epsilon| + \zeta_\epsilon)^{-2}.
\]
Moreover, Algorithm \ref{alg:FB_alg} is $\epsilon$-LDP.
\end{theorem}
A detailed proof of the theorem is provided in Appendix~\ref{appendix:proof_FB_alg}. Note again, by using the $PrivBern(\epsilon)$ Mechanism, Lemma~\ref{lem:privbern} establishes that the Bernoulli mechanism is \(\epsilon\)-differentially private, and using the definition of an LDP-bandit algorithm as stated in Definition~\ref{def:LDPBandit}, we can immediately conclude that the Algorithm \ref{alg:FB_alg} is \(\epsilon\)-differentially private.

From Theorem \ref{thm:FB_alg}, we see that Algorithm \ref{alg:FB_alg} has expected loss that decreases exponentially as the budget $T$ increases. Note that the form of this upper bound is similar to the non-private results seen in \citep{locatelli16}, except we have a different complexity term $H_\epsilon$ instead of the usual $H$. Specifically, this adjusted term accounts for the effect of privacy and is given by the relation
\[
H_\epsilon = \left( \frac{e^\epsilon - 1}{e^\epsilon + 1} \right)^{-2} H,
\]
which shows how the complexity increases as a function of the privacy parameter $\epsilon$. In Section \ref{sec:FB_lb}, we show that this complexity term characterizes the difficulty of the private fixed budget thresholding bandits problem. We also show that the upper bound in Theorem \ref{thm:FB_alg} is near-optimal (minimax optimal) for small $\epsilon$ while guaranteeing $\epsilon$-LDP.

\subsection{Lower Bound for the Fixed Budget Case}
\label{sec:FB_lb}
In this section, we derive a lower bound for the performance of \(\epsilon\)-differentially private algorithms in the the fixed budget thresholding bandit problem. Our analysis focuses on algorithms interacting with \(\epsilon\)-differentially private responses. While the algorithm in the previous section was analyzed under the Bernoulli mechanism, the lower bound presented here applies to any \(\epsilon\)-differentially private mechanism. The main result of this section is summarized in the following theorem, which establishes a minimax lower bound on the expected loss for any \(\epsilon\)-LDP bandit algorithm.

\begin{theorem}\label{thm:FB_lb}
Let $K, T > 0$. For each $1 \leq i \leq K$, let $\mu_i \in [0,1]$. Let $\tau \in [0,1], \zeta \geq 0$, and $\epsilon \geq 0$.

We write $\mathcal{B}^i$ for the environment where the distribution of arm $j \in \{1, \ldots, K\}$ is $\text{Bernoulli}\left(\tau + \frac{|\mu_j-\tau|}{2} + \zeta\right)$ if $i \neq j$ and $\text{Bernoulli}\left(\tau - \frac{|\mu_j-\tau|}{2} - \zeta\right)$ if $i = j$. 

It holds that for any $\epsilon$-LDP bandit algorithm $\pi$
\[
\max_{i \in \{0, \ldots, K\}} \mathbb{E}_{\mathcal{B}^i} \left[ \mathcal{L}(T) \right] \geq \frac{1}{4} \exp \left( -\frac{8T}{H_\epsilon} (e^\epsilon + 1)^2 \min \{4, e^{2\epsilon} \} \right),
\]
where $\mathbb{E}_{\mathcal{B}^i}$ is the expectation w.r.t. the measure induced by the interactions between policy $\pi$ and environment $\mathcal{B}^i$.
\end{theorem}

Comparing the lower bound here in Theorem \ref{thm:FB_lb} and the upper bound from Theorem \ref{thm:FB_alg}, we see that Algorithm \ref{alg:FB_alg} is near-optimal when $\epsilon \to 0$, i.e. when maximal level of privacy is achieved. In particular, this is because, as $\epsilon \rightarrow 0$, 
$(e^\epsilon + 1)^2 \min \{4, e^{2\epsilon}\} = O(1). $
Hence the upper and lower bounds match up to constants and an additive $\log\log T$ term. A detailed proof of Theorem \ref{thm:FB_lb} is provided in Appendix~\ref{appendix:proof_FB_lb}.

\section{Fixed Confidence Case}
\subsection{Algorithm for the Fixed Confidence Case}
\label{sec:FC_alg}

In this section, we propose a private fixed confidence thresholding bandit algorithm, motivated by the non-private approach studied by \cite{Chen14}. Again several adjustments need to be made to the algorithm to account for the fact that we only observe the private responses, not the true rewards.

The algorithm begins by pulling each arm once and collecting private responses through the $PrivBern(\epsilon)$ mechanism. 
At each subsequent round, Algorithm \ref{alg:FC_alg} selects the set of arms whose empirical means exceed a threshold which has been adjusted to account for privacy. It then computes a confidence radius for each arm.
If the threshold is within the confidence interval for an arm's mean, that is, if it is still uncertain whether the arm is above or below the threshold, this is a potential arm to explore. The algorithm plays the arm with the largest uncertainty among the ones whose confidence interval intersects the threshold, and repeats the process. Otherwise, it finalizes the classification of each arm.

We theoretically analyze the performance of Algorithm \ref{alg:FC_alg} and provide an upper bound for the sample complexity in Theorem \ref{thm:FC_alg}. We demonstrate that, with high probability, it correctly identifies the optimal set of arms, with the total number of exploration rounds bounded by \( O(H_\epsilon \log(\frac{4K H_\epsilon}{\delta})) \). Here, \( H_\epsilon \) captures the cumulative difficulty of distinguishing arms above the threshold based on their privatized rewards. Moreover, we show that Algorithm~\ref{alg:FC_alg} is $\epsilon$-LDP.

\begin{algorithm}

\caption{Fixed confidence thresholding LDP-bandit algorithm}\label{alg:FC_alg}

\begin{algorithmic}[1]

\State \textbf{Input:} $\tau$, $\delta \in (0, 1)$, $\epsilon \geq0$

\State Pull each arm once and receive private responses from $PrivBern(\epsilon)$, $t \gets K$, $T_i(t) \gets 1 \,\, \forall i \in [K]$

\State $\hat{\mu}_i^t :=$ empirical mean of the private responses of arm $i$ till time $t$

\For{$t = K, K+1, \dots$}

    \State $S_t \gets \{i \in [K] \mid \hat{\mu}_i^t \geq \tau_{\epsilon}\}$

    \State Compute confidence radius $\text{rad}(i) \,\, \forall i \in [K]$
    \Comment where
    \(
    \text{rad}_t(i) = \sqrt{\frac{\log\left(\frac{4 K t^3}{\delta}\right)}{8T_i(t)}}
    \)
    \For{$i = 1, \dots, K$}

        \If{$i \in S_t$}

            \State $\tilde{\mu}_i^t \gets \hat{\mu}_i^t - \text{rad}_t(i)$

        \Else

            \State $\tilde{\mu}_i^t \gets \hat{\mu}_i^t + \text{rad}_t(i)$

        \EndIf

    \EndFor

    \State $\tilde{S}_t \gets \{i \in [K] \mid \tilde{\mu}_i^t \geq \tau_{\epsilon}\}$

    \If{$S_t = \tilde{S}_t$}

        \State \Return $S_t$ 

    \EndIf

    \State $I_t \gets \arg\max_{i \in (S_t \setminus \tilde{S}_t) \cup (\tilde{S}_t \setminus S_t)} \text{rad}_t(i)$

    \State Pull arm $I_t$ and observe the private response from $PrivBern(\epsilon)$

    \State Update empirical means $\hat{\mu}_{I_t}^t$ using the private reward

    \State Update number of pulls: $T_{I_t}(t+1) \gets T_{I_t}(t) + 1$

\EndFor

\end{algorithmic}

\end{algorithm}

 \begin{theorem}\label{thm:FC_alg}

Let $K > 0$, and let all the arms rewards be in $[0,1]$. Let $\tau \in \mathbb{R}$, $\epsilon \geq 0$, $\delta \in (0, 1)$. Then, with probability at least $1 - \delta$, algorithm \ref{alg:FC_alg} returns the optimal set and

\[
T \leq O\left(H_\epsilon \log\left(\frac{4K H_\epsilon}{\delta}\right)\right).
\]

Moreover, Algorithm \ref{alg:FC_alg} is $\epsilon$-LDP.
\end{theorem}
A detailed proof of the theorem is provided in Appendix~\ref{appendix:proof_FC_alg}.

Note that, in the fixed confidence case, we set the tolerance parameter $\zeta_\epsilon = 0$, and thus  
\[
H_\epsilon = \sum_{i=1}^K (\mu_{i,\epsilon} - \tau_\epsilon)^{-2} =\left( \frac{e^\epsilon - 1}{e^\epsilon + 1} \right)^{-2} H.
\]

We emphasize that the complexity term, $H_\epsilon$, that appears in our upper bound is different to the complexity term in upper bounds attained in the non-private setting, $H$. Once again, we show that this upper bound for the sample complexity is near-optimal (in an instance-dependent sense)  for small $\epsilon$ in Section \ref{sec:FC_lb}. This demonstrates that $H_\epsilon$ quantifies the complexity of ensuring $\epsilon$-LDP in the fixed confidence thresholding bandits problem.

\subsection{Lower Bound for the Fixed Confidence Case}
\label{sec:FC_lb}
In this section, we establish a lower bound on the expected stopping time \( T \) for any \(\epsilon\)-LDP bandit algorithm that guarantees a correct solution with \(\delta\)-confidence. This means that after \( T \) iterations, the set of arms returned by the algorithm will be correct with probability greater than \( 1 - \delta \). The setup involves an environment where each arm's reward is modeled as a Bernoulli random variable with mean \(\mu_i\), and the objective is to identify arms whose mean rewards exceed a predefined threshold applying an \(\epsilon\)-differentially private mechanism to generated rewards. The proof builds upon lower bound results for non-private bandit algorithms, specifically those presented in \citep[Appendix C, Theorem 1]{Chelsire20} and \citep[Appendix B.2, Theorem 2.1]{Victor24}.

\begin{theorem}\label{thm:FC_lb}
Let $K \geq 0$, $\delta \in (0,1)$ and $\epsilon \geq 0
$. Assume that, for any $ 0 <i \leq K$, $\mu_i \in [0,1]$. Let $Q$ be any such environment with Bernoulli distributed rewards. 
It holds that for any $\epsilon$-LDP bandit algorithm
\[
\E{Q}{T} \geq  \Omega \left( H_\epsilon \log \left(\frac{1}{\delta}\right) \left( 2 \min\{4, e^{2\epsilon}\} (e^\epsilon + 1)^2 \right)^{-1} \right)
\]
where $T$ is the stopping time of the algorithm, and the set of arms returned is correct with probability at least $1 - \delta$.
\end{theorem}
Note that by comparing the lower bound in Theorem \ref{thm:FC_lb} and the upper bound in Theorem \ref{thm:FC_alg}, we see that as $\epsilon \to 0$ Algorithm \ref{alg:FC_alg} is instance-dependent optimal up to constants and additive $\log$ terms. 

A detailed proof of the theorem is provided in Appendix~\ref{appendix:proof_FC_lb}. 

\section{Conclusion}
\label{sec:conclusion}
In this work, we investigated the thresholding bandit problem under local differential privacy constraints, proposing and analyzing algorithms for both fixed budget and fixed confidence settings. In the fixed budget case, our analysis provided near-matching (up to poly-logarithmic factors) upper and lower bounds on the expected loss for small $\epsilon$ (high privacy levels), showing the optimality of our method. In the fixed confidence setting, we also provided near-matching upper and lower bounds for the sample complexity for high privacy regimes.  
Our results highlight the fundamental trade-off between privacy and sample efficiency in thresholding bandits, showing the relationship between the privacy level $\epsilon$ and the complexity of the two problem settings. While differential privacy introduces additional noise, our findings show that effective algorithmic adjustments can mitigate the performance degradation. 
Future work could explore other privacy-preserving frameworks or consider extending other pure exploration bandit problems to the private setting. Another avenue of future work could study low privacy regimes. For now, we have filled a gap in the private bandits literature by providing optimal algorithms for the private fixed confidence and fixed budget thresholding bandits problems under high privacy regimes.

\section*{Acknowledgments}
Annalisa Barbara acknowledges support from the EPSRC Centre for Doctoral Training in Statistics and Machine Learning EP/Y034813/1 and from the MUR - PRIN 2022 project 2022R45NBB funded by the Next Generation EU program.

\bibliography{references}

\newpage
\section*{Appendix}
\appendix
\section{Proof of Lemma~\ref{lem:privbern}}\label{appendix:proof_privbern}

\begin{proof}
Let arm $a$ with mean reward $\mu_a$ and privacy parameter $\epsilon > 0$ be given. Let $R$ denote the reward returned by pulling the arm $a$ and use $B(R)$ to denote the output of $PrivBern(\epsilon)$ observed by the agent. The value of $B$ is either $1$ or $0$.

\medskip

\noindent Let $r, r' \in [0, 1]$ be given. Observe that
\[
\mathbb{P}\{B(r) = 1\} = \frac{r e^\epsilon + 1 - r}{e^\epsilon + 1} = \frac{1}{2} + (2r - 1) \cdot \frac{e^\epsilon - 1}{2(e^\epsilon + 1)},
\]
and since $\mathbb{E}[R] = \mu_a$, the mean reward of arm $a$, we obtain
\[
\mathbb{E}[B(R)]  = \mathbb{E}\left[\frac{1}{2} + (2R - 1) \cdot \frac{e^\epsilon - 1}{2(e^\epsilon + 1)}\right] = \frac{1}{2} + (2\mu_a - 1) \cdot \frac{e^\epsilon - 1}{2(e^\epsilon + 1)}.
\]
This proves the mean of the returned value.

\medskip

\noindent Since $\mathbb{P}\{B(r) = 1\}$ is increasing with respect to $r$, we have
\[
\frac{\mathbb{P}\{B(r) = 1\}}{\mathbb{P}\{B(r') = 1\}} \leq \frac{\mathbb{P}\{B(1) = 1\}}{\mathbb{P}\{B(0) = 1\}} = \frac{\frac{e^\epsilon}{e^\epsilon + 1}}{\frac{1}{e^\epsilon + 1}} = e^\epsilon.
\]
Also, since $\mathbb{P}\{M(r) = 0\}$ is decreasing in $r$, we have
\[
\frac{\mathbb{P}\{
B(r) = 0\}}{\mathbb{P}\{B(r') = 0\}} \leq \frac{\mathbb{P}\{B(1) = 0\}}{\mathbb{P}\{B(0) = 0\}} = \frac{\frac{1}{e^\epsilon + 1}}{\frac{1}{e^\epsilon + 1}} = e^\epsilon.
\]

\noindent Thus, we conclude that  the mechanism $PrivBern$ is $\epsilon$-DP.
\end{proof}

\section{Proof of Theorem~\ref{thm:FB_alg}}\label{appendix:proof_FB_alg}
\begin{proof}
We first notice that by Lemma~\ref{lem:privbern}, it follows immediately that algorithm \ref{alg:FB_alg} is $\epsilon$-LDP.

We will now show that on a well-chosen event $\Lambda$, we correctly classify the arms which have mean reward larger than 
$\tau_\epsilon + \zeta_\epsilon$ and reject the arms that are under $\tau_\epsilon - \zeta_\epsilon$.

\textbf{A favorable event.}
Let $\delta = (4\sqrt{2})^{-1}$. We define the event $\Lambda$ as follows:
\[
\Lambda = \left\{ \forall i \in A, \forall s \in \{1, \dots, T\} : 
\left|\frac{1}{s} \sum_{t=1}^s X_{i,t} - \mu_{i,\epsilon}\right| \leq \sqrt{\frac{T \delta^2}{H_\epsilon s}} \right\}
\]
where \(X_{i,t}\) denotes the privatized reward observed by the agent after pulling arm $i$ at time $t$.

By the Sub-Gaussian martingale inequality, and noting that Bernoulli random variables are \(\frac{1}{4}\)-subgaussian, we have the following for each \(i \in A\) and \(u \in \{0, \dots, \lfloor\log(T)\rfloor\}\):
\[
\mathbb{P}\left( \exists v \in [2^u, 2^{u+1}], \left\{ \left|\frac{1}{v} \sum_{t=1}^v X_{i,t} - \mu_{i,\epsilon}\right| \geq \sqrt{\frac{T \delta^2}{H_\epsilon v}} \right\} \right) \leq \exp\left(-\frac{8T \delta^2}{ H_\epsilon}\right).
\]

$\Lambda$ is the union of these events for all $i \leq K$ and $s \leq \lfloor\log(T)\rfloor$. As there are less than $(\log(T) + 1)K$ such combinations, we can lower-bound its probability of occurrence with a union bound by:
\[
\mathbb{P}(\Lambda) \geq 1 - 2(\log(T) + 1)K \exp\left(-\frac{8T \delta^2}{H_\epsilon}\right).
\]

\textbf{Characterization of some helpful arm.}
At time $T$, we consider an arm $k$ that has been pulled after the initialization phase and such that $T_k(T) - 1 \geq \frac{(T-K)}{H_\epsilon \Delta_{k, \epsilon}^2}$. We know that such an arm exists; otherwise, we get:
\[
T - K = \sum_{i=1}^K (T_i(T) - 1) < \sum_{i=1}^K \frac{T-K}{H_\epsilon \Delta_{i, \epsilon}^2}= T-K
\]
which is a contradiction. Note that since $T \geq 2K$,  as specified in the theorem, we have that $T_k(T) - 1 \geq \frac{T}{2H_\epsilon \Delta_{i, \epsilon}^2}$.

We now consider $t^* \leq T$, the last time this arm $k$ was pulled. Using $T_k(t^*) \geq 2$, we know that:
\begin{equation}\label{eq:pulls_lowbound}
T_k(t^*) \geq T_k(T) - 1 \geq \frac{T}{2H_\epsilon \Delta_{k, \epsilon}^2}.
\end{equation}

\textbf{Lower bound on the number of pulls of the other arms.}
On $\Lambda$, at time $t^*$, we have for every arm $i$:
\begin{equation}\label{eq:priv_concentration}
|\hat{\mu}_i(t^*) - \mu_{i, \epsilon}| \leq \sqrt{\frac{T \delta^2}{H_\epsilon T_i(t^*)}}. 
\end{equation}

From the reverse triangle inequality we have:
\begin{align*}
|\hat{\mu}_i(t^*) - \mu_{i,\epsilon}| &= |\hat{\mu}_i(t^*) - \tau_\epsilon - (\mu_{i,\epsilon} - \tau_\epsilon)| 
 \geq ||\hat{\mu}_i(t^*) - \tau_\epsilon| - |\mu_{i,\epsilon} - \tau_\epsilon||
\\ &\geq |(|\hat{\mu}_i(t^*) - \tau_\epsilon| + \zeta_\epsilon) - (|\mu_{i,\epsilon} - \tau_\epsilon| + \zeta_\epsilon)|
\geq |\hat{\Delta}_{i,\epsilon}(t^*) - \Delta_{i,\epsilon}|.
\end{align*}

Combining this with equation~\eqref{eq:priv_concentration}, we get the following:
\begin{equation}\label{eq:Delta_bound}
\Delta_{k, \epsilon}(t^*) - \sqrt{\frac{T \delta^2}{H_\epsilon T_k(t^*)}}  \leq \hat{\Delta}_{k,\epsilon}(t^*)\leq \Delta_{k,\epsilon} + \sqrt{\frac{T \delta^2}{H_\epsilon T_k(t^*)}}. 
\end{equation}

By construction, we know that at time \( t^* \) we pulled arm \( k \), which yields for every arm \( i\):
\begin{equation}\label{eq:B_bound}
B_k(t^*) \leq B_i(t^*).
\end{equation}

We can lower bound the left-hand side of~\eqref{eq:B_bound} using equation~\eqref{eq:pulls_lowbound}:

 \begin{align}
     B_k(t^*) &\geq \left( \Delta_{k,\epsilon} - \sqrt{\frac{T \delta^2}{H_\epsilon T_k(t^*)}} \right) \sqrt{T_k(t^*)}\nonumber\\
     &\geq \left( \Delta_{k, \epsilon} - \sqrt{2} \delta \Delta_k \right) \sqrt{\frac{T}{2 H_\epsilon \Delta_k^2}}\nonumber\\
     &\geq \left( \frac{1}{\sqrt{2}} - \delta \right) \sqrt{\frac{T}{H_\epsilon}} \leq B_k(t^*),\label{eq:B_bound1}
 \end{align}

and upper bound the right-hand side using~\eqref{eq:Delta_bound} by:
\begin{align}
B_i(t^*) &= \hat{\Delta}_{i,\epsilon} \sqrt{T_i(t^*)}\nonumber\\
&\leq \left( \Delta_{i,\epsilon} + \sqrt{\frac{T \delta^2}{H_\epsilon T_i^2}} \right) \sqrt{T_i(t^*)}\nonumber\\
&\leq \Delta_{i,\epsilon} \sqrt{T_i(t^*)} + \delta \sqrt{\frac{T}{H_\epsilon}}, \label{eq:B_bound2}
\end{align}

As both \(\hat{\Delta}_{i,\epsilon}\) and \(\Delta_i\) are positive by definition, combining~\eqref{eq:B_bound1}  and~\eqref{eq:B_bound2} yields the following lower bound on the event \( T_i(T) \geq T_i(t^*) \):
\begin{equation}\label{eq:T_bound}
\left( 1 - 2 \sqrt{2 \delta} \right)^2 \frac{T}{2 H_\epsilon \Delta_{i,\epsilon}^2} \leq T_i(t^*).
\end{equation}

\textbf{Conclusion.} On \( \Lambda \), as \(\Delta_{i,\epsilon}\) is a positive quantity, combining ~\eqref{eq:priv_concentration} and ~\eqref{eq:T_bound} yields:
\begin{equation}\label{eq:mu_bound}
\mu_{i,\epsilon} - \Delta_{i,\epsilon}\frac{ \sqrt{2}\delta}{1 - 2 \sqrt{2 \delta}} \leq \hat{\mu}_i(T) \leq \mu_{i,\epsilon} + \Delta_{i,\epsilon} \frac{\sqrt{2}\delta}{1 - 2 \sqrt{2 \delta}},
\end{equation}

where \(\frac{\sqrt{2}\delta}{1 - 2 \sqrt{2 \delta}}\) simplifies to \(1/2\) for \(\delta = (4 \sqrt{2})^{-1}\).

For arms such that \(\mu_{i,\epsilon} \geq \tau_\epsilon + \zeta_\epsilon\), then \(\Delta_{i,\epsilon} = \mu_{i,\epsilon} - \tau_\epsilon + \zeta_\epsilon\), and we can rewrite ~\eqref{eq:mu_bound} :
\begin{align}
\hat{\mu}_i(T) - \tau_\epsilon &\geq \mu_{i,\epsilon} - \tau_\epsilon - \frac{1}{2} \Delta_{i,\epsilon} \nonumber\\
&\geq (\mu_{i,\epsilon} - \tau_\epsilon) \left( 1 - \frac{1}{2} \right) - \frac{\zeta_\epsilon}{2} \nonumber\\
& \geq 0
\end{align}
where the last line uses \( \mu_{i,\epsilon} \geq \tau_\epsilon + \zeta_\epsilon \). Similarly, \( \hat{\mu}_i(T) - \tau_\epsilon < 0 \) holds for \( \mu_{i_\epsilon} < \tau_\epsilon \). On \( \Lambda \), since all the arms are correctly classified, arms with privatized mean rewards over \( \tau_\epsilon + \zeta_\epsilon \) are all accepted, and arms with privatized mean rewards under \( \tau_\epsilon - \zeta_\epsilon \) are all rejected, which means the loss suffered by the algorithm is \(0\). As \(1 - P(\Lambda) \leq 2(\log(T) + 1) K \exp \left(-\frac{4T}{H_\epsilon} \right) \), this concludes the proof.

\end{proof}

\section{Proof of Theorem~\ref{thm:FB_lb}}\label{appendix:proof_FB_lb}
\begin{proof}

Let us consider $K$ real numbers $\mu_i \in [0,1]$ and set $\tau := \frac{1}{2}, \zeta = 0$.
We write $\nu_i := \text{Bern}(\frac{1}{2} + \frac{\Delta_i}{2})$ for the Bernoulli distribution with mean $\frac{1}{2} + \frac{\Delta_i}{2}$ and $\nu_i' := \text{Bern}(\frac{1}{2} - \frac{\Delta_i}{2})$ for the Bernoulli distribution with mean $\frac{1}{2} - \frac{\Delta_i}{2}$. Note that this construction can be easily generalized to cases where $\tau \neq \frac{1}{2}$ or $\zeta \neq 0$.

We can equivalently define the environment $\mathcal{B}^i$ as the product distribution $\nu_1^i \times \ldots \times \nu_K^i$, where for $j \leq K$, $\nu^i_j := \nu_j 1_{j \neq i} + \nu_j' 1_{j = i}$.

We also extend this notation to $\mathcal{B}^0$, where $\forall j, \nu_j^0 := \nu_j$.

Let $k \in {1, \dots, K}$. For Bernoulli arms it holds that:
\begin{align}
\text{KL}(\nu_k', \nu_k) &= \left(\frac{1}{2}-\frac{\Delta_k}{2}\right)\log\frac{\left(\frac{1}{2}-\frac{\Delta_k}{2}\right)}{\left(\frac{1}{2}+\frac{\Delta_k}{2}\right)} + \left(\frac{1}{2}+\frac{\Delta_k}{2}\right)\log\frac{\left(\frac{1}{2}+\frac{\Delta_k}{2}\right)}{\left(\frac{1}{2}-\frac{\Delta_k}{2}\right)} \nonumber
\\ &=-\Delta_k \log\left(\frac{1}{2}-\frac{\Delta_k}{2}\right) +\Delta_k \log\left(\frac{1}{2}+\frac{\Delta_k}{2}\right) \nonumber
\\&\leq 2\Delta_k^2, \label{eq:KL_bernoulli}
\end{align}
where the last inequality follows since $\log(1+x)\leq x$ for $x\geq -1$.

A lowerbound on $\max_{i \in \{0, \ldots, K\}}\mathbb{E}_{\mathcal{B}^i}\left[\mathcal{L}_i(T)\right]$
can be derived as follows. Let $\hat{S}$ denote the set of arms returned by the algorithm, and let $S_i^*$ represent the correct set of arms in the environment $\mathcal{B}^i$.

\begin{align}
\max_{i \in \{0, \ldots, K\}} \mathbb{E}_{\mathcal{B}^i}\left[\mathcal{L}_i(T)\right] &= \max_{i \in \{0, \ldots, K\}} \mathbb{P}_i(\hat{S} \neq S_i^*) \nonumber\\
&\geq \frac{1}{2} \left( \mathbb{P}_i(\hat{S} \neq S_i^*) + \mathbb{P}_0(\hat{S} \neq S_0^*) \right) \label{eq:max_set}\\
&\geq \frac{1}{2} \left( \mathbb{P}_i(\hat{S} \neq S_i^*) + \mathbb{P}_0(\hat{S} = S_i^*) \right) \label{eq:events}\\
&\geq \frac{1}{4} \exp\left(-\text{KL}(\mathbb{P}_i, \mathbb{P}_0)\right) \quad \label{eq:bretagnolle}.
\end{align}

where ~\eqref{eq:max_set} holds since the maximum  of a set is larger than an average of a subset, ~\eqref{eq:events} holds since the event $\hat{S} \neq S_0^*$ includes the event $\hat{S}=S_i^*$ (where $i$ is chosen arbitrarily) and ~\eqref{eq:bretagnolle} holds by Bretagnolle-Huber  inequality.

From the locally private KL divergence decomposition as given in \cite[Lemma 4]{basu2019privacy}, we obtain
\begin{align}
\frac{1}{4} \exp\left(-\text{KL}(\mathbb{P}_i, \mathbb{P}_0)\right)
&\geq \frac{1}{4} \exp \left( -\sum_{j=1}^K \mathbb{E}_{\mathbb{P}_i}\left[T_j(T)\right] \cdot 2\min\{4, e^{2\epsilon}\}(e^\epsilon - 1)^2\text{KL}_j(\mathbb{P}_i, \mathbb{P}_0) \right) \nonumber\\
&\geq \frac{1}{4} \exp \left( -\sum_{j=1}^K \mathbb{E}_{\mathbb{P}_i}\left[T_j(T)\right] \cdot 2\min\{4, e^{2\epsilon}\}(e^\epsilon - 1)^2 2 \Delta_j^2  1_{j = i} \right) \label{eq:bernoulli}\\
&\geq \frac{1}{4} \exp \left( -\mathbb{E}_{\mathbb{P}_i}\left[T_i(T)\right] \cdot 2 \Delta_i^2 \cdot 2\min\{4, e^{2\epsilon}\}(e^\epsilon - 1)^2 \right)\nonumber
\end{align}
where $\text{KL}_j(\mathbb{P}_i, \mathbb{P}_0)$ denotes the KL divergence between the distributions of arm $j$ in setting $\mathcal{B}^i$ and $\mathcal{B}^0$ respectively, inequality ~\eqref{eq:bernoulli} follows from  ~\eqref{eq:KL_bernoulli} and $1_{j = i}$ is an indicator function taking value equal to 1 if $j=i$ and 0 otherwise. 

Since $\sum T_i(T) = T$ and all $T_i(T)$ are positive, there exists an arm $i$ such that
\(
T_i(T) \leq \frac{4T}{H \Delta_i^2}.
\)
Otherwise $T=\sum_i T_i(T) > \sum_i \frac{4T}{H\Delta_i^2}= T$, which is a contradiction.

Let $i$ be an arm for which \(T_i(T) \leq \frac{4T}{H \Delta_i^2}.\) For arm $i$ we have
\begin{align}
\frac{1}{4} \exp \left( -\mathbb{E}_{\mathbb{P}_i}\left[T_i(T)\right] \cdot 2 \Delta_i^2 \cdot 2\min\{4, e^{2\epsilon}\}(e^\epsilon - 1)^2 \right)
&\geq \frac{1}{4} \exp\left(-\frac{4T}{H \Delta_i^2} \cdot 2 \Delta_i^2 \cdot 2\min\left\{4, e^{2\epsilon}\right\}(e^\epsilon - 1)^2\right) \nonumber \\
&= \frac{1}{4} \exp\left(-\frac{8T}{H} \cdot 2\min\left\{4, e^{2\epsilon}\right\}(e^\epsilon - 1)^2\right) \nonumber\\
&= \frac{1}{4} \exp\left(-\frac{8T}{H_\epsilon} (e^\epsilon + 1)^2 \cdot 2\min\left\{4, e^{2\epsilon}\right\}\right) \label{eq:H_substitution}
\end{align}
where equation  ~\eqref{eq:H_substitution} is obtained by replacing $H=\frac{(e^\epsilon-1)^2H_\epsilon}{(e^\epsilon+1)^2}$.
\end{proof}

\section{Proof of Theorem~\ref{thm:FC_alg}}\label{appendix:proof_FC_alg}
\begin{proof}
We first notice that by Lemma~\ref{lem:privbern}, it follows immediately that algorithm \ref{alg:FC_alg} is $\epsilon$-LDP. 

We now prove the bound on \( T \) as stated in the theorem. Let \( S^* \) denote the correct solution, i.e., the set of arms with mean reward above the threshold \( \tau \). Our goal is to show that after \(O\left(H_\epsilon \log\left(\frac{4K H_\epsilon}{\delta}\right)\right)\) iterations, the algorithm returns the correct set with probability greater than \( 1 - \delta \).

    \textbf{Correctness under a favorable event.} Denote the event
    \[
    A_t = \{\forall i \in [K] \mid |\mu_{i,\epsilon} - \hat{\mu}_i^t| \leq \text{rad}_t(i)\}
    \]
    that occurs when the confidence bounds of all arms are valid at round $t$.
    
    If algorithm \ref{alg:FC_alg} terminates at time $t$, none of the confidence intervals intersect with $\tau_\epsilon$. Specifically, the confidence interval is:
    \begin{itemize}
    \item $[\hat{\mu}_i^t - \text{rad}_t(i), \hat{\mu}_i^t]$ if $i \in S_t$, i.e., if the empirical mean is above the threshold,
    \item $[\hat{\mu}_i^t, \hat{\mu}_i^t + \text{rad}_t(i)]$ if $i \notin S_t$, i.e., if the empirical mean is below the threshold.
    \end{itemize}
    In fact, if at time $t$ there is an arm $i$ whose empirical mean is at distance less than $\text{rad}_t(i)$ from $\tau_\epsilon$, then $S_t \neq \tilde{S}_t$ and so the algorithm could not terminate at iteration $t$. Moreover, if $A_t$ holds, all the confidence intervals are valid. This implies that 
    \[\mu_{i, \epsilon} \geq \tau_\epsilon \Rightarrow i \in S_t\]
    and 
    \[\mu_{i, \epsilon} <\tau_\epsilon \Rightarrow i \notin S_t.\]
    
    Since, by definition, $\mu_{i, \epsilon} \geq \tau_\epsilon \iff \mu_i \geq \tau$, it then  follows that 
    \[
    \mu_i \geq \tau \Rightarrow i \in S_t
    \]
    and
    \[
    \mu_i < \tau \Rightarrow i \notin S_t.
    \]
    Hence given any $t \geq K$, if $A_t$ holds and algorithm \ref{alg:FC_alg} terminates at round $t$, we can conclude that $S_t=S^*$ where $S^*$ denotes the correct solution.

    \textbf{High-probability guarantee.}
    We define the event \( A \) as the simultaneous occurrence of the events \( A_t \) at all time steps \( t \), that is,
    \[
    A := \bigcap_{t=1}^{\infty} A_t.
    \]
    We begin by showing that \( A \) holds with high probability:

    \[
    \mathbb{P}[A] = \mathbb{P}\left[\bigcap_{t=1}^{\infty} A_t \right] \geq 1 - \delta.
    \]
    In fact,
    \begin{align*}
    \mathbb{P}\left[\bigcap_{t=1}^{\infty} A_t \right]&= 1 - \mathbb{P}\left[\overline{\bigcap_{t=1}^{\infty} A_t} \right]\\ &= 1- \mathbb{P}\left[\bigcup_{t=1}^{\infty} \overline{A_t} \right]\\ &\geq 1 - \sum_{t=1}^\infty  \mathbb{P}\left[ \overline{A_t} \right]\\ &\geq 1-\sum_{t=1}^\infty \frac{\delta}{2t^2}\\ &=1- \frac{\pi^2 \delta}{12} \geq 1- \delta
     \end{align*}
    where the second inequality holds since by fixing any $t>0$ and $i \in [n]$, by Hoeffding inequality, we have
    \begin{align*}
    \mathbb{P}\left[|\hat{\mu}_i^t-\mu_{i,\epsilon}| \geq \frac{1}{4} \sqrt{\frac{2\log\left(\frac{4 K t^3}{\delta}\right)}{T_i(t_i)}}\right]&= \sum_{s=1}^{t-1} \mathbb{P}\left[|\hat{\mu}_i^t-\mu_{i,\epsilon}| \geq \frac{1}{4} \sqrt{\frac{2\log\left(\frac{4 K t^3}{\delta}\right)}{s}}\Bigg|T_i(t)=s\right] \\ &\leq \sum_{s=1}^{t-1} \frac{\delta}{2Kt^3}\\ &\leq \frac{\delta}{2Kt^2}
    \end{align*}
  The last inequality provides a bound on \( \mathbb{P}\left[ \overline{A_t} \right] \), which, when substituted into the previous expression, yields a \( 1 - \delta \) bound.
    
    \item \textbf{Characterization of arm selection.} The next step is to prove that for any $t > 0$, if $A_t$ occurs and
    \(
    \text{rad}_t(i) \leq \frac{\Delta_i}{2},
    \)
    then arm $i$ will not be pulled at round $t$.

    To prove this, suppose that \( \text{rad}_t(i) \leq \frac{\Delta_{i,\epsilon}}{2}\) and \(A_t\) holds. Then
    \[
    \hat{\mu}_{i, \epsilon} - \frac{1}{2}\Delta_{i, \epsilon} \leq \mu_{i, \epsilon} \leq \hat{\mu}_{i, \epsilon} + \frac{1}{2} \Delta_{i, \epsilon}.
    \]
    \begin{itemize}
        \item Case 1 ($\mu_{i, \epsilon} \geq \tau_\epsilon$): $\hat{\mu}_{i, \epsilon} \geq \mu_{i, \epsilon}- \text{rad}_t(i)\Rightarrow \hat{\mu}_{i,\epsilon} - \text{rad}_t(i) \geq \mu_{i, \epsilon}- 2\text{rad}_t(i) \geq \mu_{i, \epsilon}-\Delta_{i, \epsilon} = \tau_\epsilon$\\
        implying that arm $i$ will not be pulled at time $t$ since $i \notin (S_t \setminus \tilde{S}_t) \cup (\tilde{S}_t \setminus S_t)$.
        \item Case 2 ($\mu_{i, \epsilon} < \tau_\epsilon$): $\hat{\mu}_{i, \epsilon} \leq \mu_{i, \epsilon}+ \text{rad}_t(i) \Rightarrow \hat{\mu}_{i,\epsilon} + \text{rad}_t(i) \leq \mu_{i, \epsilon}+ 2\text{rad}_t(i) \leq \mu_{i, \epsilon}+\Delta_{i, \epsilon} = \tau_\epsilon$\\
        implying that arm $i$ will not be pulled at time $t$ since $i \notin (S_t \setminus \tilde{S}_t) \cup (\tilde{S}_t \setminus S_t)$.
    \end{itemize}
    \textbf{Conclusion.} Denote by $t_i$ the last time arm $i$ is pulled and by $T(i)$ the total number of pulls of arm $i$. Then, $T_i(t_i) = T(i) - 1$. By point 3, we know $\text{rad}_{t_i}(i) > \frac{\Delta_{i, \epsilon}}{2}$, and by substituting \(
    \text{rad}_t(i) = \sqrt{\frac{\log\left(\frac{4 K t^3}{\delta}\right)}{8T_i(t)}}
    \), we get

    \[
    \frac{\Delta_{i,\epsilon}}{2} \leq \frac{1}{4}\sqrt{\frac{2\log\left(\frac{4 K t^3}{\delta}\right)}{T_i(t_i)}} \leq \frac{1}{4}\sqrt{\frac{2\log\left(\frac{4 K T^3}{\delta}\right)}{T(i) - 1}}.
    \]

    Isolating $T(i)$ from the above we obtain:

    \[
    T(i) \leq \frac{1}{2\Delta_{i, \epsilon}^2} \log\left(\frac{4 K T^3}{\delta}\right) + 1.
    \]

    Now define $\tilde{H}_\epsilon := \max \{\frac{1}{36} H_\epsilon, 1 \}$. In the rest of the proof we show that 
    \begin{equation}\label{eq:FC_bound}
    T \leq 499 \tilde{H}_\epsilon \log\left(\frac{4K \tilde{H}_\epsilon}{\delta}\right)+ 2K 
     \end{equation}
    If $K \geq \frac{1}{2}T$, then we see that $T \leq 2K$ and therefore equation ~\eqref{eq:FC_bound} holds immediately.\\
    If $K < \frac{1}{2}T$ (and so $T>K)$, we can write that 
    \begin{equation}\label{eq:T_def}
    T =C \tilde{H}_\epsilon \log\left(\frac{4K \tilde{H}_\epsilon}{\delta}\right)+ K.
    \end{equation}
    If $C\leq 499$ then equation ~\eqref{eq:FC_bound} holds, while if $C>499$ we obtain the following:
    \begin{align}
    T &\leq K + \sum_{i \in [K]} \frac{1}{2\Delta_{i,\epsilon}^2} \log\left(\frac{4KT^3}{\delta}\right) \nonumber \\
    &\leq K + 8 \tilde{H}_\epsilon \log\left(\frac{4KT^3}{\delta}\right)
    \nonumber\\
    &= K + 8 \tilde{H}_\epsilon \log\left(\frac{4K}{\delta}\right) + 24 \tilde{H}_\epsilon \log(T) \nonumber\\
    &\leq K + 8 \tilde{H}_\epsilon \log\left(\frac{4K}{\delta}\right) + 24 \tilde{H}_\epsilon \log\left(2C \tilde{H}_\epsilon \log\left(\frac{4K \tilde{H}_\epsilon}{\delta}\right)\right)\label{eq:proof_eq1} \\   &= K + 8 \tilde{H}_\epsilon \log\left(\frac{4K}{\delta}\right) + 24 \tilde{H}_\epsilon \log(2C) + 24 \tilde{H}_\epsilon \log \tilde{H}_\epsilon + 24 \tilde{H}_\epsilon \log\log\left(\frac{4K \tilde{H}_\epsilon}{\delta}\right)\nonumber\\
    &\leq K + 8 \tilde{H}_\epsilon \log\left(\frac{4K \tilde{H}_\epsilon}{\delta}\right) + 24 \tilde{H}_\epsilon \log(2C)\log\left(\frac{4K \tilde{H}_\epsilon}{\delta}\right) + 24 \tilde{H}_\epsilon \log\left(\frac{4K \tilde{H}_\epsilon}{\delta}\right) + 24 \tilde{H}_\epsilon \log\left(\frac{4K \tilde{H}_\epsilon}{\delta}\right)\label{eq:proof_eq2}\\
    &\leq K + \left(56 + 24 \log(2C)\right) \tilde{H}_\epsilon \log\left(\frac{4K \tilde{H}_\epsilon}{\delta}\right)\label{eq:proof_eq3}\\
    &< K + C \tilde{H}_\epsilon \log\left(\frac{4K \tilde{H}_\epsilon}{\delta}\right) \quad \text{for some } C > 499\label{eq:proof_eq4}\\
    &=T \label{eq:proof_eq5}
    \end{align}
    where Equation ~\eqref{eq:proof_eq1} follows from Equation ~\eqref{eq:T_def} and the assumption that $K < \frac{1}{2}T$; Equation ~\eqref{eq:proof_eq2} follows from the fact that $\tilde{H}_\epsilon \geq 1$ and $\delta <1$; Equation ~\eqref{eq:proof_eq4} follows since $56+24\log(2C) < C$ for all $C >499$; and Equation ~\eqref{eq:proof_eq5} is due to Equation ~\eqref{eq:T_def}. Now we see that ~\eqref{eq:proof_eq5} is a contradiction. Therefore we obtain that $C\leq 499$ and we have proved Equation  ~\eqref{eq:FC_bound}.

\end{proof}

\section{Proof of Theorem~\ref{thm:FC_lb}}\label{appendix:proof_FC_lb}
\begin{proof}
Let \(\nu\) represent an arbitrary environment, and let \(S_\nu\) denote the set of arms with mean rewards greater than a threshold \(\tau\). Suppose \(\pi\) is a policy that, after stopping at time \(T\), returns the correct set of arms with probability at least \(1-\delta\) for any environment \(\nu\). That is, for all environments \(\nu\),

\[
\forall \nu \quad \Prob_{\pi_\nu} \left( \hat{S}_T = S_\nu\right) \geq 1-\delta.
\]

Furthermore, assume that \(\pi\) is \(\epsilon\)-LDP.

Now, consider an arbitrary environment Q and let \(\left(\mu_{Q,i}\right)_{i=1}^{K}\) represent the set of mean rewards for \(K\) arms in Q, where \(\mu_{Q,i} \in (0,1)\) for each \(i\).

Now, we define a family of environments \(Q^j\) for each $j\in \{1,\dots,K\}$ as follows:

\begin{itemize}
    \item \textbf{Case 1: \(\mu_{Q,j} < \tau\)}. For the \(j\)-th arm, define the mean reward in environment \(Q^j\) as
    \[
    \mu_{Q^j,j} = \text{clip}_{[0,1]}\left(\tau + \left|\tau - \mu_{Q,j}\right| \right),
    \]
    and for all \(i \neq j\),
    \[
    \mu_{Q^j,i} = \mu_{Q,i}.
    \]

    \item \textbf{Case 2: \(\mu_{Q,j} \geq \tau\)}. For the \(j\)-th arm, define the mean reward as
    \[
    \mu_{Q^j,j} = \text{clip}_{[0,1]}\left(\tau - \left|\tau - \mu_{Q,j}\right| \right),
    \]

\end{itemize}
    and for all \(i \neq j\),
    \[
    \mu_{Q^j,i} = \mu_{Q,i}.
    \]
Here, the \(\text{clip}_{[0,1]}\) function is defined as:

\[
\text{clip}_{[0,1]}(x) =
\begin{cases}
x & \text{if } 0 \leq x \leq 1, \\
1 & \text{if } x > 1, \\
0 & \text{if } x < 0.
\end{cases}
\]

By Lemma 1 in  \citep{kaufmann2016complexity}, we have the following bound for the Kullback-Leibler divergence:

\[
\KL{\Prob_{\pi, Q}}{\Prob_{\pi, Q^i}} \geq k\ell\left(\Prob_{\pi, Q}(\hat{S} = S_{Q}), \Prob_{\pi, Q^i}(\hat{S} = S_{Q})\right).
\]

Using the KL-decomposition lemma for \(\epsilon\)-differentially private mechanisms \citep{basu2019privacy}, we upper bound the left-hand side by \(\DPconst \Delta^2 \mathbb{E}_{Q}[T_i(T)]\), and by applying the inequality \( k\ell(x,y) \geq x\log\left(\frac{1}{y}\right) - \log(2) \) for \(x,y \in [0,1]\), we obtain a lower bound for the right-hand side:

\[
\DPconst \Delta^2 \mathbb{E}_{Q}[T_i(T)] \geq \Prob_{\pi, Q}(\hat{S} = S_{Q}) \log \left(\frac{1}{\Prob_{\pi, Q^i}(\hat{S} = S_{Q})}\right) - \log(2).
\]

Since we assume the policy always returns the correct set of arms with probability greater than \(1 - \delta\), and the set of correct arms are different under $Q$ and $Q^i$, we have:

\[
\Prob_{\pi, Q}(\hat{S} = S_{Q}) \geq 1 - \delta \quad \text{and} \quad \Prob_{\pi, Q^i}(\hat{S} = S_{Q}) \leq \delta.
\]

Thus, we get the following inequality:

\[
\DPconst \Delta^2 \mathbb{E}_{Q}[T_i(T)] \geq (1 - \delta) \log \left(\frac{1}{\delta}\right) - \log(2).
\]

Rearranging terms, we obtain:

\[
\mathbb{E}_{Q}[T_i(T)] \geq \frac{1}{\DPconst \Delta^2} \left( (1 - \delta) \log \left(\frac{1}{\delta}\right) - \log(2) \right).
\]

Finally, summing over all \(K\) arms, we get the total expected regret:

\begin{align*}
\mathbb{E}_{Q}[T] = \sum_{i=1}^{K} \mathbb{E}_{Q}[T_i(T)] 
&\geq \sum_{i=1}^{K} \frac{1}{\DPconst \Delta^2} \left( (1 - \delta) \log \left(\frac{1}{\delta}\right) - \log(2) \right) \\
&= \frac{(1 - \delta) \log \left(\frac{1}{\delta}\right) - \log(2)}{\DPconst} \sum_{i=1}^{K} \frac{1}{\Delta_i^2} \\
&= \frac{(1 - \delta) \log \left(\frac{1}{\delta}\right) - \log(2)}{\DPconst} H \\
&= \frac{(1 - \delta) \log \left(\frac{1}{\delta}\right) - \log(2)}{2 \min\{4, e^{2\epsilon}\}(e^\epsilon + 1)^2} H_\epsilon \\
&= \Omega \left( H_\epsilon \log \left(\frac{1}{\delta}\right) \left( 2 \min\{4, e^{2\epsilon}\} (e^\epsilon + 1)^2 \right)^{-1} \right).
\end{align*}
\end{proof}

\end{document}